\documentclass{article}

\usepackage[preprint,nonatbib]{neurips_2020}

\usepackage[utf8]{inputenc} 
\usepackage[T1]{fontenc}    
\usepackage{hyperref}       
\usepackage{url}            
\usepackage{booktabs}       
\usepackage{amsfonts}       
\usepackage{nicefrac}       
\usepackage{microtype}      
\usepackage{booktabs} 
\usepackage{url}
\usepackage{float}
\usepackage{subfig}
\usepackage{graphicx}
\usepackage[noend]{algorithmic}
\usepackage{algorithm}
\usepackage{multirow}
\usepackage{amsmath}
\usepackage{amsthm}
\usepackage{xcolor}
\usepackage{wrapfig}

\title{Deep Neural Network for 3D Surface Segmentation based on Contour Tree Hierarchy}

\author{%
   Wenchong He \\
   The University of Alabama \\
   Tuscaloosa, AL, 35487 \\
   \texttt{whe11@crimson.ua.edu} \\
   \AND
   Arpan Man Sainju \\
   The University of Alabama \\
   Tuscaloosa, AL, 35487 \\
   \texttt{asainju@crimson.ua.edu} \\
   \And
   Zhe Jiang \\
   The University of Alabama \\
   Tuscaloosa, AL, 35487 \\
   \texttt{zjiang@cs.ua.edu}
   \And
   Da Yan \\
   The University of Alabama at Birmingham \\
   Birmingham, AL, 35294 \\
   \texttt{yanda@uab.edu}
}

\begin{document}

\maketitle

\begin{abstract}
Given a 3D surface defined by an elevation function on a 2D grid as well as non-spatial features observed at each pixel, the problem of surface segmentation aims to classify pixels into contiguous classes based on both non-spatial features and surface topology. The problem has important applications in hydrology, planetary science, and biochemistry but is uniquely challenging for several reasons. First, the spatial extent of class segments follows surface contours in the topological space, regardless of their spatial shapes and directions. Second, the topological structure exists in multiple spatial scales based on different surface resolutions. Existing widely successful deep learning models for image segmentation are often not applicable due to their reliance on convolution and pooling operations to learn regular structural patterns on a grid. In contrast, we propose to represent surface topological structure by a contour tree skeleton, which is a polytree capturing the evolution of surface contours at different elevation levels. We further design a graph neural network based on the contour tree hierarchy to model surface topological structure at different spatial scales.  Experimental evaluations based on real-world hydrological datasets show that our model outperforms several baseline methods in classification accuracy.
\end{abstract}

\section{Introduction}\label{sec:intro}
Given a 3D surface defined by an elevation function over a 2D grid as well as non-spatial features observed at each pixel, the problem of surface segmentation aims to classify pixels into segment classes based on non-spatial features and surface topology. For example, in hydrology, scientists are interested in mapping the surface water extent from remote sensing imagery based on not only spectral features of pixels but also the geographic terrains. The problem is unique from traditional image segmentation in that the spatial extent of class segments follow a topological structure based on surface contour patterns. For example, flood extent boundary spreads over a geographic terrain with equal elevation contours.

{\em Societal applications:} Topological surface segmentation is important in many applications such as flood extent mapping on a topographic surface in hydrology, crater detection in planetary surface, molecular surface segmentation in bio-science~\cite{natarajan2006segmenting},  plant branching structure analysis in phenology. In hydrology, the topological surface is a digital elevation map, the non-spatial features can be optical remote sensing imagery imposed on the surface, and the target output map can be water surface extent. Mapping water surface extent plays an important role in national water forecasting and disaster response. In astronomy, scientists are interested in detecting craters on a 3D planetary surface based on the topological structures. Such information can reveal details on how a planet evolves over time. In bio-science, people are interested in segmenting protein surface based on the topology to identify rigid components and to understand the function of cavities and protrusions in protein-protein interactions~\cite{natarajan2006segmenting}. In botany, researchers are interested in assessing and comparing the branching structure of plants based on the surface topology. Such structures are important features to understand the phenotype of a plant (the interactions of genetic background with the environment).

However, the problem poses several unique challenges. First, the spatial extent of class segments follows surface contours in the topological space, regardless of their spatial shapes and directions. This violates the assumption by the widely popular deep convolutional neural networks that the spatial structure of class segments is regular on a Euclidean plane (a grid framework). Second, the topological structure exists in multiple spatial scales based on different surface resolutions. Specifically, on a fine resolution, a surface will show more local fluctuations; but on a coarse resolution, a surface will only capture the global trend ignoring local details. 

Existing techniques for 3D surface segmentation can be categorized into traditional non-deep learning approaches and deep learning approaches. Non-deep learning approaches include region growing, hierarchical clustering, iterative clustering, and graph algorithms~\cite{shamir2008survey,nguyen20133d}. The region growing approach~\cite{kalvin1996superfaces} starts with one or several seed elements and greedily expands the seed into regions by adding in other likely elements. Among the region's growing methods, one important technique is called the watershed method~\cite{grau2004improved}, which grows seeds based on a height function defined on the surface (e.g., from a local minimum to a ridge). The hierarchical clustering approach differs from the region growing approach in that it considers all vertices as individual seeds (clusters) and greedily selects two small clusters to merge into a bigger cluster in each iteration~\cite{garland2001hierarchical}. The iterative clustering algorithm (e.g., K means) can iteratively assign vertices into different clusters based on objective criteria. The graph-based approach considers the surface as a graph and uses graph-cut algorithms to segment the surface~\cite{chen2012three}. Deep learning has achieved great progress for image segmentation~\cite{minaee2020image}. The most popular technique is to learn a fully convolutional network~\cite{ronneberger2015u,long2015fully} to extract high-level semantic features together with deconvolution or upsampling layers to combine features at multiple scales for detailed segmentation~\cite{chen2017deeplab,chen2017rethinking,noh2015learning}. These methods often require partitioning the image into smaller patches (e.g., 224 by 224) and learning a fully convolutional network for each patch with the depth channel as an additional feature~\cite{wang2016learning,ma2017multi,hazirbas2016fusenet}, without explicitly modeling the topological constraint. Some works aim to resolve this issue by adding spatial transformation in convolution kernels through a depth or Gaussian term, e.g., bilateral filters~\cite{barron2016fast} and depth-aware CNN~\cite{wang2018depth}, but they still do not fully capture the topological structure for the entire surface. 
Other methods incorporate topological constraints into image segmentation in two ways: enforcing MRF or CRF-based topological constraints in the inference step~\cite{vicente2008graph,nowozin2009global,chen2011enforcing,andres2011probabilistic}, which is unable to fully utilize a topological prior to train a model; or using a topology-aware loss function to train the neural network by leveraging persistent homology to define a topological loss on the predicted class image~\cite{hu2019topology,mosinska2018beyond}, which only focuses on basic topology constraint (e.g., the number of connected components or holes). One example is retinal layer segmentation in medical application~\cite{he2019deep,lang2017improving}, which only focus on learning contiguous boundaries between layers.  In summary, existing methods do not fully incorporate the topological structures in the form of surface contours.

In contrast, we propose a novel graph neural network model for 3D surface segmentation by representing surface topological structure as a \emph{contour tree skeleton}. A contour tree is a polytree capturing the evolution of surface contours at different elevation levels in computational topology. We further design a graph neural network based on the \emph{contour tree hierarchy} to model surface topological structure at different spatial scales. We design downsampling and upsampling paths in the graph neural network based on the contour tree hierarchy to extract features at different scales. Experimental evaluations based on real-world hydrological datasets show that our model outperforms several baseline methods in classification accuracy. 

\section{Problem Statement}\label{sec:prob}
{\bf 3D surface:} A 3D surface is defined as an elevation (or depth) function on a 2D grid. We denote the elevation function by an array $\mathbf{E}\in \mathbb{R}^{N\times N}$, where $N$ is the number of pixels in each spatial dimension. On a 3D surface, there can be $m$ non-spatial explanatory features as well as a target class layer. We denote each pixel as $\mathbf{s}_n=(\mathbf{x}_n, e_n, y_n)$, where $\mathbf{x}_n$, $e_n$, and $y_n$ represent the $m$ explanatory features, elevation value, and the class of the pixel respectively. Examples of 3D surface include the geographic terrains on the Earth's surface based on digital elevation and local protein elevation surface in biochemistry. Note that in these cases, the underlying 2D grid is actually on a sphere, but it can be considered as a flat in a small local area.

Given a 3D surface with non-spatial explanatory features and target classes, the problem aims to learn a model to predict surface classes based on the explanatory features and the surface topological structure. For example, in hydrology, scientists are interested in mapping the surface water extent from remote sensing imagery based on not only the spectral signature of pixels but also the geographic terrain in a digital elevation map.

\section{The Proposed Approach}\label{sec:approach}
This section introduces our proposed approach to address the unique challenges in topological surface segmentation. The key idea is to represent a topological surface by a contour tree skeleton. We design a graph neural network model with graph convolutional operations on contour tree nodes. In order to learn a topological structure at different spatial scales, we construct a contour-tree hierarchy and design downsampling and upsampling paths in the graph neural network. 

\subsection{Overview of deep model architecture}
\begin{figure}
    \centering
    \includegraphics[width=1 \textwidth]{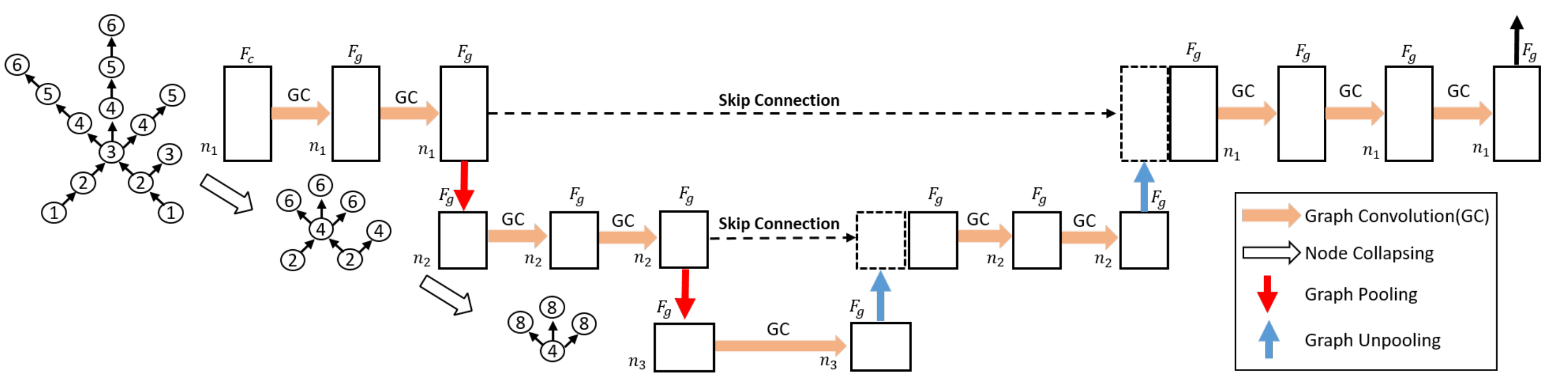}
    \caption{Overview of the model architecture}
    \label{fig:overview}
\end{figure}
Figure~\ref{fig:overview} provides an overview of the proposed model. The entire model architecture looks similar to the U-Net model~\cite{ronneberger2015u} in traditional 2D image segmentation. The input of the model is a contour tree with extracted features on each contour node (through aggregating the features of individual pixels on the same contour). The node features are illustrated by a rectangle whose height represents the number of nodes and whose width represents the feature dimension. Two consecutive graph convolution operations are added at each level to further learn features on the topological skeleton. In the downsample path on the left, pooling operation is added by aggregating contour tree nodes based on node collapsing operations. In the upsample path on the right, unpooling operations are used to project features to a fine resolution. Pooled features at a coarse-scale are concatenated with features at a fine-scale from the downsample path through skip connection to combine both global and local features for detailed segmentation. 

\subsection{Surface representation by a contour tree}
The key idea in the proposed approach is to represent the {\bf topological skeleton} of a surface by a contour tree, which is a fundamental concept in computational topology. We first introduce several important concepts and formally define a contour tree.

\begin{wrapfigure}{r}{0.5 \textwidth}
    \centering
    \subfloat[]{\includegraphics[height=0.9in]{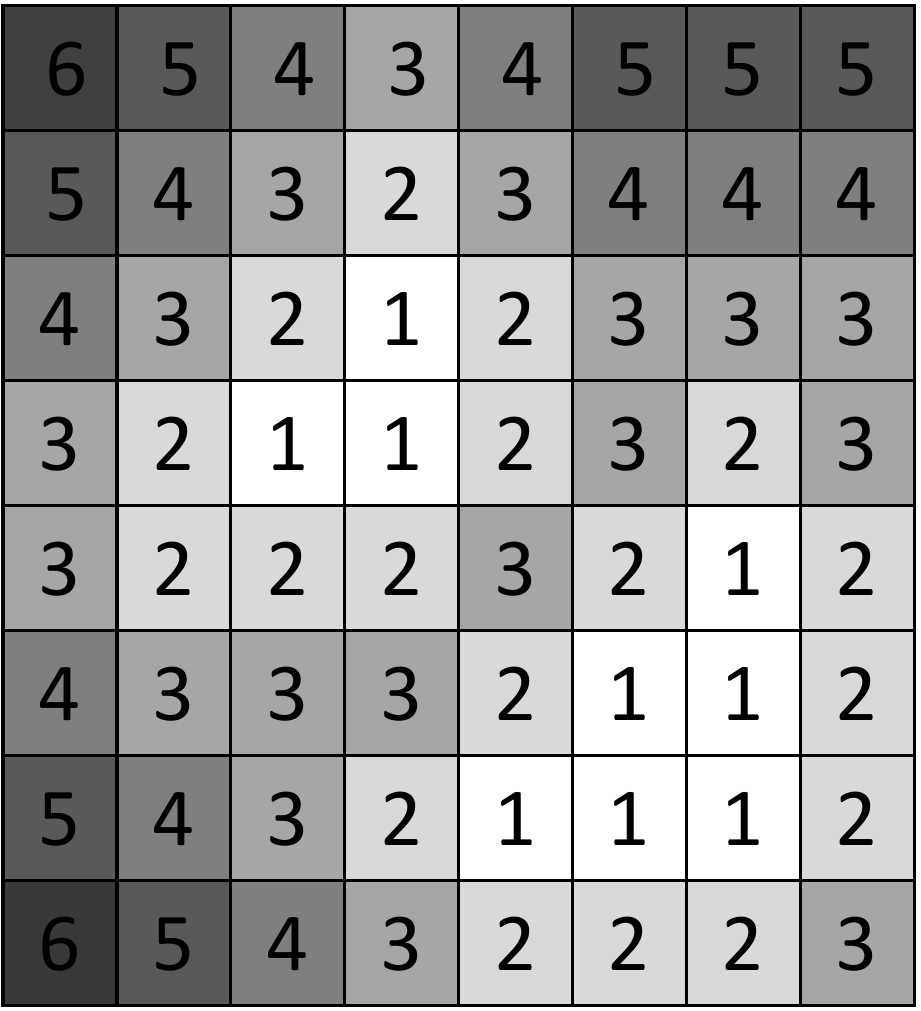}}
    \hspace{0.5mm}
    \subfloat[]{\includegraphics[height=0.9in]{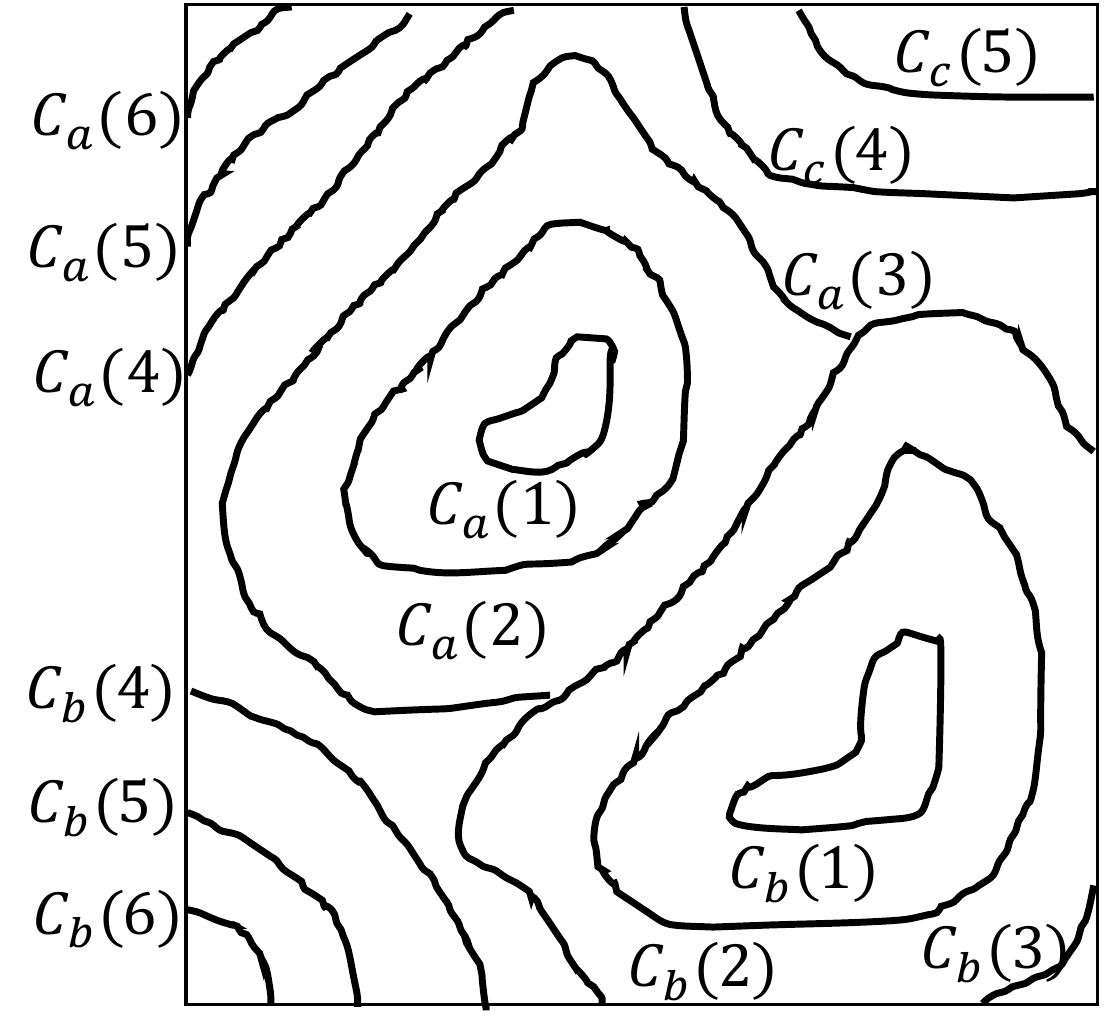}}
    \hspace{0.5mm}
    \subfloat[]{\includegraphics[height=0.8in]{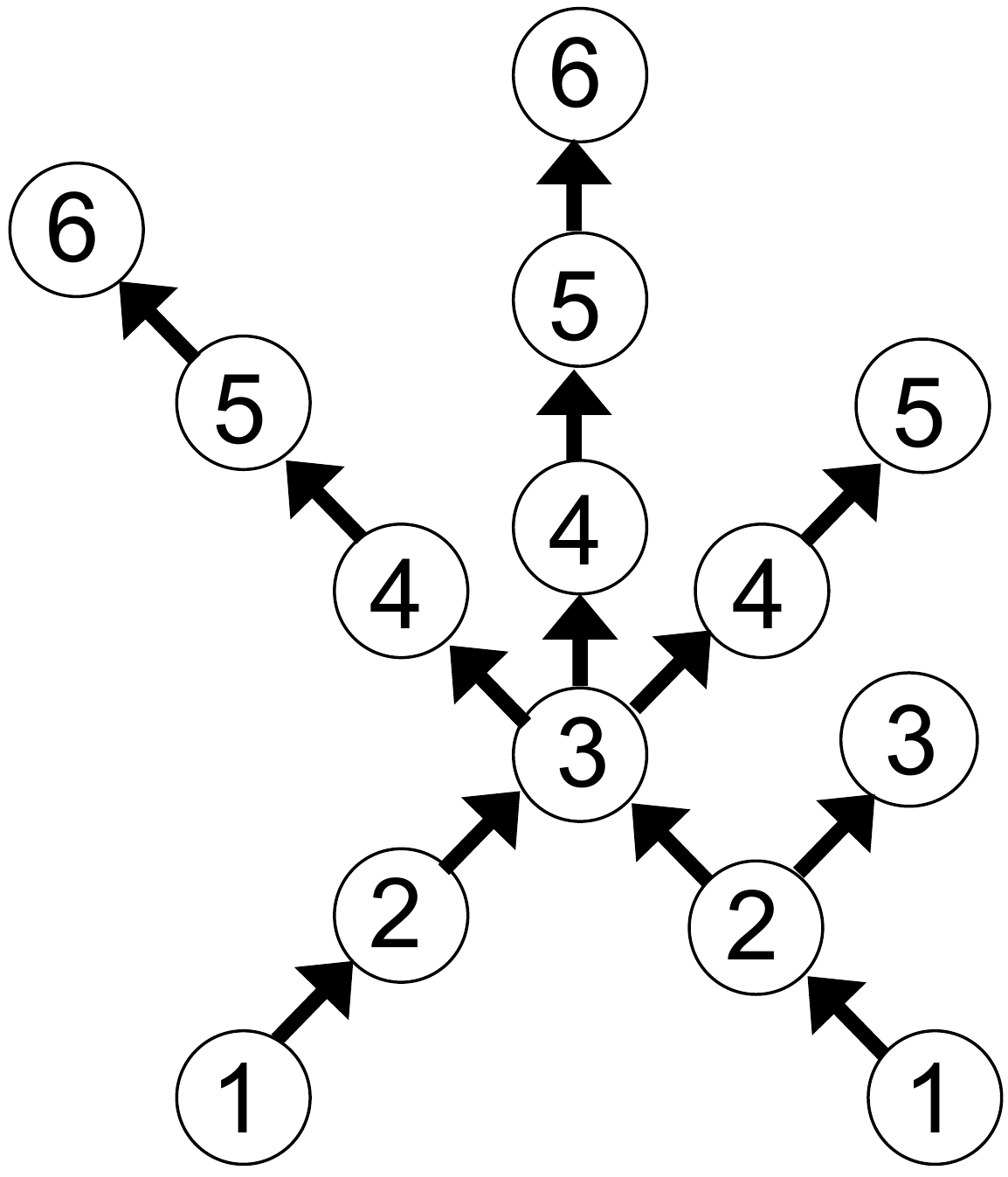}}
    \caption{An example of an elevation surface (a), its contours (b) and contour tree (c)}
    \label{fig:ct}
    \vspace{-3mm}
\end{wrapfigure}

{\bf Level set, contour:} A level set on a 3D surface is the set of pixels with an equal elevation. Formally, $L(e_0)=\{\mathbf{s}_n|e_n=e_0\}$, where $e_0$ is an elevation threshold. Specifically, each level set on a 3D surface is composed of one or several connected components called contours, i.e., $L(e_0)=\bigcup_{k=1}^K C_k(e_0)$, where $C_k(e_0)$ is only self-connected (no connectivity between contours). For example, in a 3D surface defined by the elevation function in Figure~\ref{fig:ct}(a), there are fourteen separate contour components in total. For instance, there are two contours $C_{a}(1)$ and $C_{b}(1)$ for elevation $1$, as well as three contours $C_{a}(4)$ and $C_{b}(4)$ for elevation $4$, shown in Figure~\ref{fig:ct}(b).

{\bf Topological structure:} In this paper, we define topological structure as the evolution of the topology of surface contours along with elevation levels. Specifically, as the elevation level rises, new contours can appear, merge, split, and disappear. Consider the example of Figure~\ref{fig:ct}(a). 
As the elevation level rises from $1$ to $6$, two contours $C_{a}(1)$ and $C_{b}(1)$ first appear, then spread to $C_{a}(2)$ and $C_{b}(2)$ respectively, merge into $C_{a}(3)$ for elevation $3$, split into three contours $C_{a}(4),C_{b}(4),C_{c}(4),C_{d}(4)$, and finally disappear after elevation $6$. The topological structure provides a constraint on how the spatial extent of class segments spread over the surface. Pixels on the same contour have the same segment class. If we look at the evolution of a contour, we can find a topological order such as $C_{a}(1)\rightarrow C_{a}(2)\rightarrow C_{a}(3)$. For example, in hydrological applications, the spatial extent of water on the surface in Figure~\ref{fig:ct}(a) is constrained by the contour patterns. 

{\bf Contour tree:} A contour tree is a polytree whose nodes are surface contours and whose edges are the topological order (based on elevation gradient directions) between two adjacent contours. Formally, a contour tree can be denoted by $\mathcal{T}=(\mathcal{V},\mathcal{E})$, where $\mathcal{V}$ is the set of contours denoted by 
$\mathcal{V}=\{C_k(e)|e_{min}\leq e \leq e_{max}\}$ ($e_{min}$ and $e_{max}$ are the minimum and maximum elevation), and $\mathcal{E}\subset\mathcal{V}\times \mathcal{V}$ are the topological order between adjacent contours based on the elevation gradient. Contour tree is a fundamental tool to represent a surface topological structure in computational topology.
For example, the contour tree corresponding to the 3D surface in Figure~\ref{fig:ct}(a) is shown in Figure~\ref{fig:ct}(c), where the tree branches show the evolution of contours at different elevation levels. The contour tree structure provides a topological structural constraint on how a potential class segment evolves on the surface. Since pixels on the same contour have the same segment class based on the topological constraint, we can aggregate features and classes from individual pixels to their corresponding contour tree nodes. We denote the node features and classes in a contour tree as $\mathbf{X}\in \mathbb{R}^{n\times m}$ and $\mathbf{Y} \in  \mathbb{R}^{n\times 1}$ respectively, where $n$ is the total number of nodes in a contour tree.

Its more general definition in the field of topology is {\em Reeb graph}~\cite{edelsbrunner2010computational}, which captures how the level sets of a smooth and differential function evolves on a manifold (more details can be found in Morse theory~\cite{munkres2000topology}). A contour tree is a special case of Reeb graph when the function is defined on a 2D plane such that its level sets are in the form of loops called contours. 

Representing the topological skeleton of a surface using a contour tree is important for our surface segmentation problem. The contour tree provides a global topological structure of a surface that often reflect the physical constraints of the target surface class segments. In other words, the contour tree structure provides an opportunity to learn structural features on a surface in the topological space. For example, in hydrology, the extent of water distribution on an elevation surface (also called geographic terrain surface) follows the topological structure of contours due to gravity. In botany, the branching structure of a planet also follows the topology of the plant body along with the height. Such topological structural features are very hard, if possible, to learn in the original surface representation (2D grid map) due to a lack of a regular shape or direction. 
Extensive research has been done in the field of computational topology to study efficient algorithms for contour tree construction~\cite{munkres2000topology,edelsbrunner2010computational}. 
We use the algorithm in~\cite{carr2003computing}, which involves scanning all locations by an decreasing (and increasing) order of elevation values to create a joint tree (and a split tree) and then merging the two trees together. Its overall time complexity is approximately $O(n\log n)$, where $n$ is the number of vertices~\cite{carr2003computing}. To apply the algorithm, we added random perturbation to make pixel elevation unique and then collapsed pixels on the same contour into one tree node.

\subsection{Graph convolution on a contour tree}
After representing a topological surface by a contour tree skeleton, we can conduct graph convolution operations on the contour tree.
A graph convolution layer generalizes the traditional convolutional layer from 2D images to a graph. This is non-trivial since a graph usually does not have a fixed neighborhood size. There exists extensive research on designing convolutional layers on graphs~\cite{zhou2018graph,zhang2018deep,wu2019comprehensive,chiang2019cluster,gao2018large}. Techniques can be categorized into spectral methods and spatial methods. The former utilizes graph Laplacian to aggregate features from neighboring nodes. The latter re-samples a neighborhood to a fixed size so that a traditional convolutional operator can be applied. It is worth to note that graph convolution does not consider the geometric structure of the mesh surface. But since the previous curvature filter layer already extracts geometric features at local vertices, the graph convolution layer simply focuses on incorporating spatial contiguity into extracted geometric features based on graph connectivity. 

Specifically, we propose to use two popular graph convolutional layers: ChebyNet~\cite{defferrard2016convolutional} and diffusion graph convolution~\cite{li2017diffusion}. ChebyNet is a spectral-based method that uses graph Laplacian matrix to average neighbor features into a node. It uses Chebyshev polynomial to avoid eigenvalue decomposition for the graph Laplacian. The order of the Chebyshev polynomial corresponds to the number of neighbor hops being incorporated. Specifically, the graph convolution operator in ChebyNet can be expressed as $\mathbf{X'}=\sigma(\sum g_\theta(\mathbf{L})\mathbf{X}+\mathbf{b})$, where $\mathbf{X'}$ and $\mathbf{X}$ are node feature matrices before and after the graph convolution, $g_\theta(\mathbf{L}) = \sum_0^{K-1}\theta_k\mathbf{T}_k(\hat{\mathbf{L}})$, $\mathbf{T}_k(\hat{\mathbf{L}})$ is the ChebyShev polynomial of order k, evaluated at the scaled Laplacian $\hat{\mathbf{L}}$  with $\theta_k$ as kernel weights~\cite{defferrard2016convolutional}. The diffusion graph convolution uses the indegree and outdegree normalized adjacency matrix to aggregate neighbor node features. Specifically, $\mathbf{X'}=\sigma(\sum_0^{K-1}(\theta_{k,1}(\mathbf{D}_{O}^{-1}\mathbf{W})^k + \theta_{k,2}(\mathbf{D}_{I}^{-1}\mathbf{W}^T)^k) \mathbf{X}+\mathbf{b})$, where $\mathbf{D}_{I}$ and $\mathbf{D}_{O}$ are diagonal indegree matrix and outdegree matrix, $\mathbf{W}$ is the node adjacency matrix, $\theta_{k,1}$, $\theta_{k,2}$ and $\mathbf{b}$ are kernel weights and bias parameters~\cite{li2017diffusion}. Multiplying the degree normalized adjacency matrix multiple times is equivalent to conducting the diffusion multiple hops. We can use multiple filters together to output multiple feature channels. Diffusion graph convolution defines convolutional operations on a directed graph based on two different node adjacency matrices on both directions. 



\subsection{Downsample and upsample paths based on contour tree hierarchy}
The purpose of the downsampling path is to extract semantic features of locations on the surface topology at a coarser resolution so that the extracted feature at each location has a global topological context. The task is non-trivial since the traditional max-pooling operation on 2D image data cannot be directly applied to a contour tree due to the lack of a regular grid structure. There exist some research on designing pooling operators on graphs. For example, one method is based on greedy node clustering (also called Graclus~\cite{dhillon2007weighted}), which iteratively merges two nodes with a high edge weight and low degrees. Another method called DIFFPOOL learns a cluster assignment matrix over graph nodes on top of the output of a GNN model~\cite{ying2018hierarchical}. 
Another work uses a trainable projection vector to perform k-max-pooling (i.e., selecting nodes based on the projected values from node features)~\cite{gao2019graph}. However, the existing pooling operations can not preserve the topological structures of the graph. To fill the gap, we propose to do pooling operation based on a {\bf multi-scale contour tree hierarchy}. The main intuition is to look at the surface height function at different precision levels. When the precision is high, more local fluctuations of the surface show up and the contour tree structure provides more local topological details. When the precision is low,  the surface will look smoother and the contour tree structure focuses more on global topological trends. 

Moreover, as the surface height precision keeps decreasing, previously separate contours (tree nodes) will merge together into one, creating a collapsing hierarchy. The process can be considered as a bottom-up hierarchical clustering of contour tree nodes. Figure~\ref{fig:overview} provides an example, where the original contour tree is downsampled twice. The original contour tree is based on elevation values rounded to integers, the second contour tree has elevation rounds to even integers, and the third contour tree is based on elevation values rounded to multiples of fours. 
The one to many mapping relationships between nodes on a coarse-scale contour tree to those on a fine-scale contour tree can be expressed in a pooling matrix $\mathbf{T}_l$, where $\mathbf{T}_{ij}=1$ if node $i$ at the level $l$ is collapsed into node $j$ in level $l+1$, and $\mathbf{T}_{ij}=0$ otherwise. Based on the pooling matrix, we can design an average-pooling layer on contour tree node features, as shown by $\mathbf{X}_{l+1}=\tilde{\mathbf{T}}_{l}\mathbf{X}_{l}$, where $\tilde{\mathbf{T}}_{l}$ is the row-normalized pooling matrix from level $l$ to level $l+1$, and $\mathbf{X}_{l}$ and $\mathbf{X}_{l+1}$ are node feature matrices at level $l$ and level $l+1$ respectively. This can be easily implemented based on sparse matrix multiplication. 

The purpose of the upsampling path is to concatenate global and local topological features and to project those features to a surface at a fine scale. The traditional transposed convolution cannot be applied due to two reasons. First, there is no regular window (neighborhood) structure on the contour tree as in the traditional 2D image. Second, the mapping of pixel locations from a coarser scale to a fine-scale is not as regular in a contour tree as in a traditional 2D image. Thus, we cannot easily design a pooling operation with learnable parameters by transposed weight matrix multiplication as in U-Net. To fill the gap, we propose to use simple unpooling based on the same one to many mapping between nodes in the hierarchical clustering. Specifically, we can upsample node features from a coarse contour tree to a fine-scale contour tree based on the transpose of the pooling matrix $\mathbf{T}_l^T$. The unpooling operation can be expressed as $\tilde{\mathbf{X}}_{l}=\mathbf{T}_{l}^T\tilde{\mathbf{X}}_{l+1}$, where $\mathbf{T}_{l}^T$ is the transpose of the pooling matrix from level $l$ to level $l+1$, and $\tilde{\mathbf{X}}_{l}$ is the upsampled feature matrix at level $l$ from the level $l+1$.

\section{Evaluation}\label{sec:eval}
The goal of the evaluation is to compare our proposed method with several baseline methods in classification performance on two real-world hydrological datasets. We also conducted self-comparison studies to evaluate the effect of different configurations in our model. 
Experiments were conducted on a workstation with four NVIDIA RTX 6000 GPUs. Unless specified otherwise, we used default parameters in open source tools for baseline methods. Candidate classification methods are listed below. The source codes, data samples, and more implementation details are provided in the supplementary materials.
\begin{itemize}
    \item {\bf Per-pixel classifiers}: We used random forest ({\bf RF}), maximum likelihood classifier ({\bf MLC}), and gradient boosted tree ({\bf GBM}) from R randomForest and gbm packages. 
    \item {\bf Fully convolutional network (FCN):} We used U-Net~\cite{ronneberger2015u} in Python. The model consists of ten double convolution layers with batch normalization, together with max-pooling layers and transposed convolution layers, for the downsample path and the upsample path. We used the Adam optimizer with a learning rate of $10^{-4}$ and a mini-batch size of 10. 
    \item {\bf Contour tree neural network (CTNN):} This is our proposed method. We implemented our codes in Python and Tensorflow. We used U-Net to extract pixel features by removing the last non-linear and softmax layers and aggregated pixel features as inputs into tree nodes. We used the Adam optimizer with a learning rate of $10^{-4}$ and a mini-batch size of 1.
\end{itemize}

\emph{Dataset description:} We used two flood mapping datasets from the cities of Grimesland and  Kinston in North Carolina during Hurricane Mathew in 2016. Explanatory features were red, green, blue bands in aerial imagery from NOAA National Geodetic Survey~\cite{ngs}. There are two classes: flood and dry. The digital elevation maps were from the University of North Carolina Libraries~\cite{ncsudem}. All data were resampled into 2-meter by 2-meter resolution. For the Grimland dataset, there are 80 training images, 120 validation images, and 16 test images. Each image size is 672 by 672 pixels, which are collapsed into 451584 contour tree nodes.  For the Kinston dataset, there are 132 training images, 113 validation images, and 113 test images. Each image size is 3000 by 100, totally $3\times10^5$ nodes in the contour tree. 

\emph{Evaluation Metric}: For classification performance evaluation, we used precision, recall, and F-score. We also used overall accuracy given that the two classes are relatively balanced.

\subsection{Classification Performance Comparison}\label{sec:cross}
\begin{table}[h]\small
\vspace{-5mm}
\centering
\caption{Classification on Real Dataset in Grimesland NC}
\label{tab:classificationComp1}
\begin{tabular}{ccccccc}
\toprule
\cmidrule(r){1-2}
Classifiers & Class & Precision &Recall & F & Avg. F& Accuracy\\ \midrule
 
\multirow{2}{*}{RF}&Dry&{0.68}&{0.75}&{0.71}&\multirow{2}{*}{0.73}&\multirow{2}{*}{0.73}\\ 
 &Flood&{0.79}&{0.72}&{0.75}&\\ 
 \midrule
 
 \multirow{2}{*}{MLC}&Dry&{0.69}&{0.67}&{0.68}&\multirow{2}{*}{0.72}&\multirow{2}{*}{0.73}\\ 
 &Flood&{0.75}&{0.77}&{0.76}&\\ 
 \midrule
 
 \multirow{2}{*}{GBM}&Dry&{0.68}&{0.75}&{0.71}&\multirow{2}{*}{0.73}&\multirow{2}{*}{0.73}\\ 
 &Flood&{079.}&{0.72}&{0.75}&\\ 
 \midrule
 
 
\multirow{2}{*}{U-Net}&Dry&{0.99}&{0.71}&{0.82}&\multirow{2}{*}{0.85}&\multirow{2}{*}{0.86}\\ 
 &Flood&{0.81}&{0.99}&{0.88}&\\ \midrule 
 
\multirow{2}{*}{CTNN}&Dry&{0.98}&{0.82}&{0.89}&\multirow{2}{*}{\bf 0.90}&\multirow{2}{*}{\bf 0.90}\\  &Flood&{0.84}&{0.99}&{0.91}&\\ 
\bottomrule
\end{tabular}
\end{table}

\begin{table}[h]\small
\centering
\caption{Classification on Real Dataset in Kinston NC}
\label{tab:classificationComp2}
\begin{tabular}{ccccccc}
\toprule
\cmidrule(r){1-2}
Classifiers & Class & Precision &Recall & F & Avg. F &Accuracy\\ 
\midrule 
 
\multirow{2}{*}{RF}&Dry&{0.82}&{0.73}&{0.77}&\multirow{2}{*}{0.74}&\multirow{2}{*}{0.74}\\ 
 &Flood&{0.65}&{0.76}&{0.70}&\\ 
 \midrule 
 
 \multirow{2}{*}{MLC}&Dry&{0.87}&{0.63}&{0.73}&\multirow{2}{*}{0.72}&\multirow{2}{*}{0.72}\\ 
 &Flood&{0.61}&{0.86}&{0.71}&\\ 
 \midrule 
 
 \multirow{2}{*}{GBM}&Dry&{0.82}&{0.73}&{0.77}&\multirow{2}{*}{0.74}&\multirow{2}{*}{0.74}\\ 
 &Flood&{0.65}&{0.76}&{0.70}&\\ 
 \midrule 
 
 
\multirow{2}{*}{U-Net}&Dry&{0.93}&{0.97}&{0.95}&\multirow{2}{*}{0.94}&\multirow{2}{*}{0.94}\\ 
 &Flood&{0.96}&{0.92}&{0.94}&\\ \midrule  
 
\multirow{2}{*}{CTNN}&Dry&{0.95}&{0.98}&{0.96}&\multirow{2}{*}{\bf 0.96}&\multirow{2}{*}{\bf 0.96}\\  &Flood&{0.96}&{0.94}&{0.95}&\\ 
\bottomrule 
\end{tabular}
\end{table}

We first compared different methods on their precision, recall, and F-score on the two real-world datasets. The results were summarized in Table~\ref{tab:classificationComp1} and Table~\ref{tab:classificationComp2} respectively. In this experiment, we used the default model architecture in CNTT with a four-level contour tree hierarchy based on different precision of elevation values at 0.01 meter, 0.1 meters, and 1 meter respectively. We used ChebyNet graph convolution operators at each level with the number of neighbor hops as 4 and 2 in each level, and the number of graph convolution operators as 16, 32, 64, and 128 at each level respectively. On the Grimesland dataset, random forest, maximum likelihood classifier, and gradient boosted tree achieved overall F-score around 0.73. The poor performance of per-pixel classifiers was due to class confusion from feature noise and obstacles. For example, pixels of tree canopies overlaying flood water have the same spectral features with those in dry areas. Without incorporating spatial dependency structure between pixel locations, these per-pixel classifiers cannot overcome the class confusion. U-Net performed much better with an average F-score of 0.85 due to its capability of learning complex spatial contextual features from input aerial imagery (e.g., color and texture of floodwater). Out CTNN model performed the best with an overall F-score of 0.90 (higher than the F-score of 0.85 in U-Net only). The reason was that the CTNN model can not only utilize the complex spatial contextual features extracted from U-Net but also incorporate topological structural dependency among class labels in the contour tree hierarchy. In this dataset, additional topological constraints made a significant difference because feature obstacles obscured some flood boundaries in the aerial imagery, making it difficult to delineate those boundaries in U-Net only.

Results on the Kinston dataset were summarized in Table~\ref{tab:classificationComp2}. Similar to the Grimesland dataset, we can observe that most non-spatial classifiers performed poorly with an overall F-score below 0.74. Both U-Net and CTNN performed very well in this data because the flood boundaries in this dataset were distinguishable in comparison to the Grimesland dataset. Thus, U-Net could successfully identify most flood boundaries with high accuracy. Our CTNN model performed slightly better than U-Net only, indicating that learning additional topological structure along surface contours still added some value in this case.  

\subsection{Self-comparison on model configurations}
We also conducted a sensitivity analysis to analyze the effect of different configurations in our CTNN model, such as the type of graph convolutional filters, the number of filters at each contour tree level, the number of neighbor hops in each graph convolution filter, and the number of contour tree hierarchical levels. Due to limited space, we only focused on the Grimesland dataset. Unless specified otherwise, we used the same configurations for our CTNN model as in Section~\ref{sec:cross}. Note that the metrics were directly reported on test data for sensitivity analysis. This is different from the results in Section~\ref{sec:cross}, in which we used independent validation data to select hyper-parameters.

\begin{figure}
    \centering
    \subfloat[]{\includegraphics[width=2in]{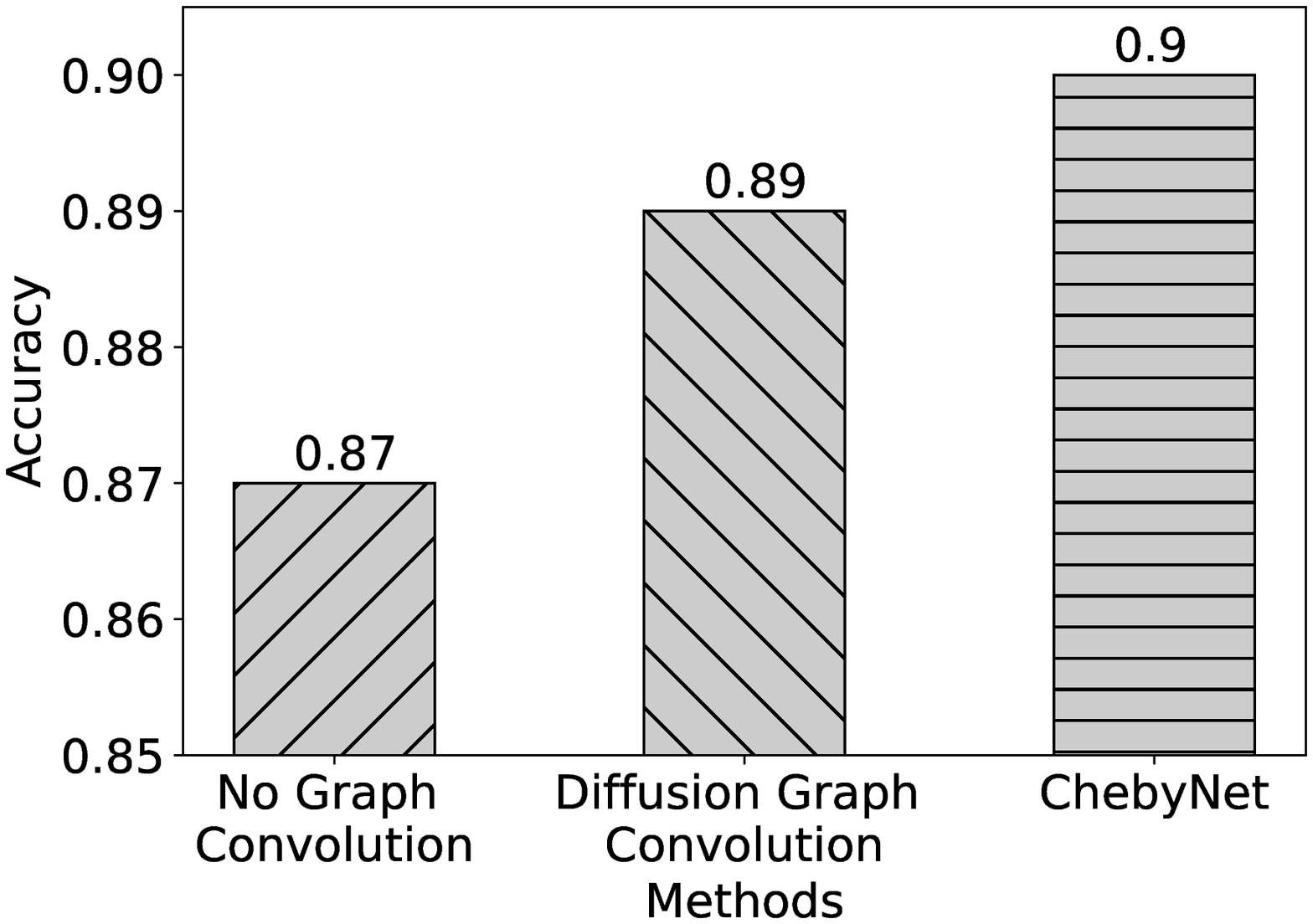}}
    \subfloat[]{\includegraphics[width=2in]{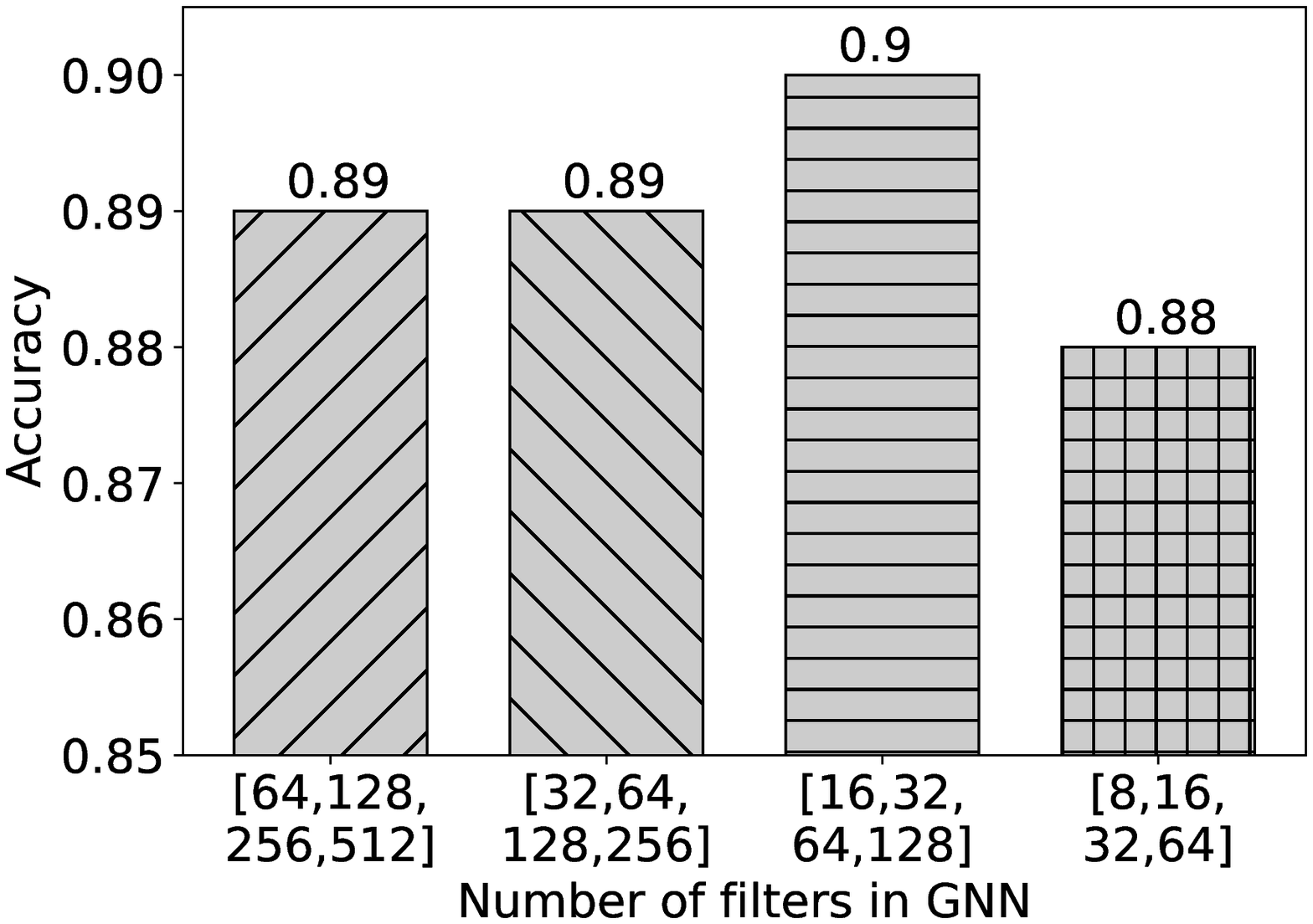}}
    \caption{The effect of graph convolutional filter types and the number of filters}
    \label{fig:selffilter}
    \vspace{-5mm}
\end{figure}

First, we studied the effect of the type of graph convolutional filters in our CTNN model. We fixed the other configurations of the model and compared three different choices: no graph convolution (i.e., only use a non-linear function to map input channels to output channels for each node, or 1D convolution), diffusion graph convolution, and ChebyNet. The results were summarized in Figure~\ref{fig:selffilter}(a). We can see that without graph convolution (i.e., only using 1D convolution on each node), the overall accuracy is 0.87. This configuration was equivalent to learning additional multi-layer perceptrons on the features of each node extracted from U-Net. Thus, the performance slightly improved over U-Net only. Adding diffusion convolution improved the overall accuracy from 0.87 to 0.89, indicating that the graph convolution filters along the contour tree structure helped better delineate flood boundaries. The accuracy of a (spectral based) ChebyNet graph convolutional filter was the highest. It was interesting that the undirected graph convolution in ChebyNet performed slightly better than the directed diffusion graph convolution on our contour tree (which is a directed graph). The reason was probably that most neighboring nodes should have the same classes due to spatial autocorrelation except for nodes near the flood boundary (the flood extent is contiguous).

\begin{figure}
    \centering
    \subfloat[]{\includegraphics[width=2in]{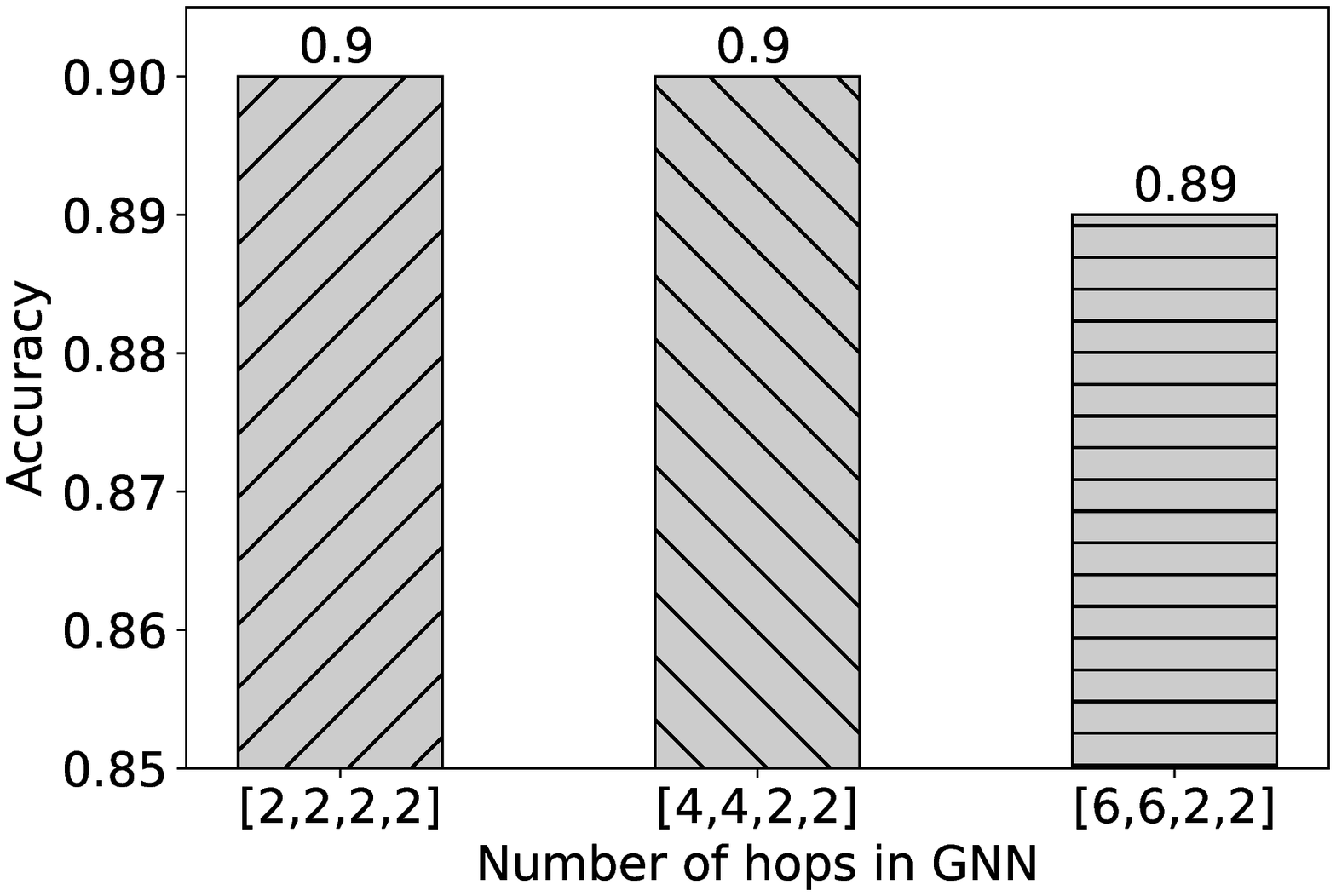}}
    \subfloat[]{\includegraphics[width=2in]{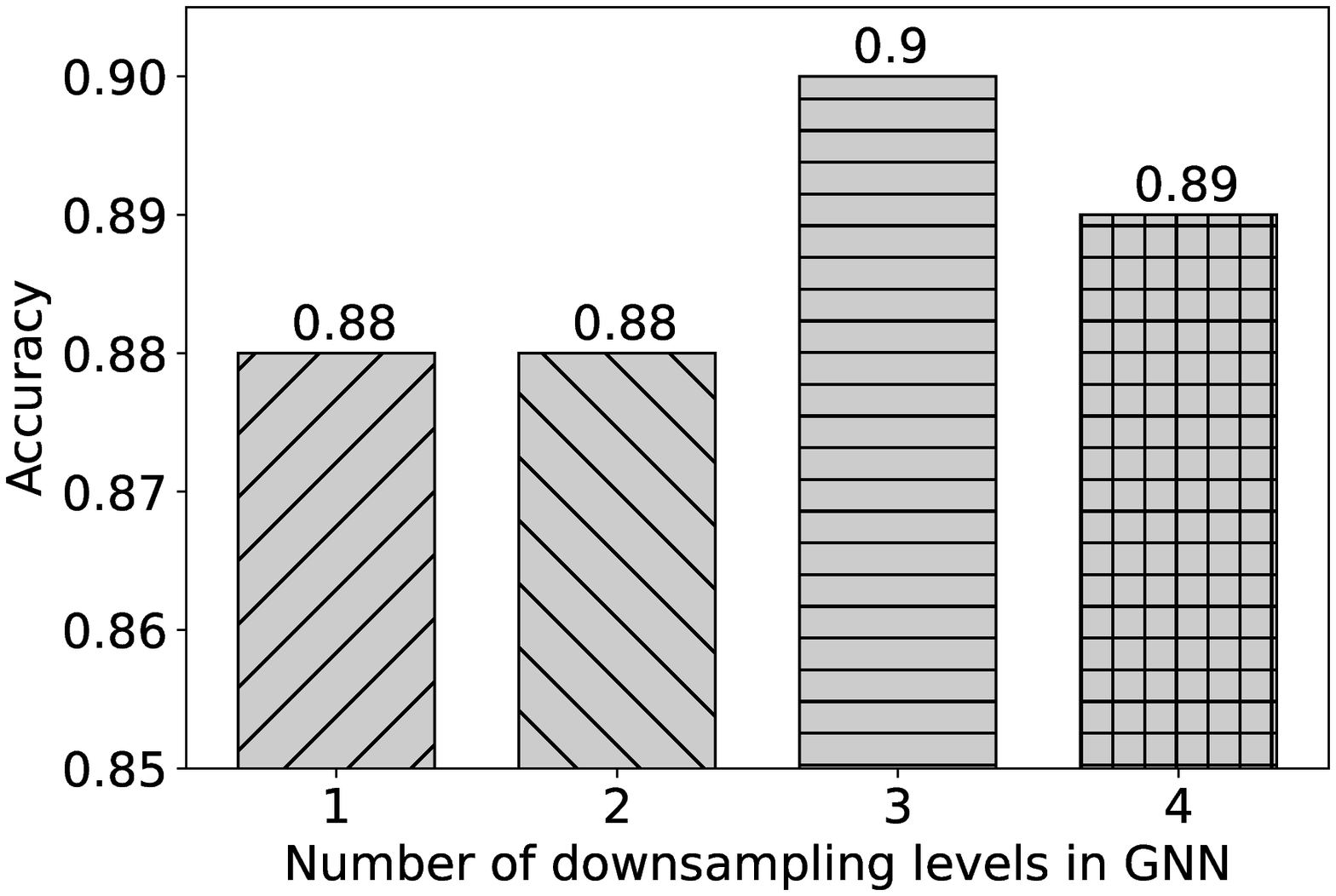}}
    \caption{The effect of the number of neighbor hops and downsample levels in GNN}
    \label{fig:self2}
\end{figure}

We also tested the effect of the number of neighbor hops in the ChebyNet graph convolutional operations. We used the default setting and varied the number of neighbor hops in three different settings. We first fixed the number of neighbor hops as $2$ in all four downsample levels, and then increased the neighbor hops of the first two downsample levels to $4$ and $6$ respectively. The comparison was summarized in Figure~\ref{fig:self2}(a). We can see that using a larger number of neighbor hops degraded the performance. This was due to the fact that a larger number of neighbor hops had an over-smoothing effect between the adjacent contours near the flood boundary. 

Finally, we also tested the effect of the number of levels in the contour tree hierarchy. We used the default setting for the other hyper-parameters and varied the number of downsampling operations from $1$ to $4$. Specifically, when using four downsample operations (five different contour tree scales), we keep reducing the precision of elevation values four times from 0.01 to 0.1 meters, 1 meter, and 10 meters respectively. We found out that downsampling the contour tree $3$ times (into four levels) had the best overall accuracy.


\section{Conclusion and Future Work}
In this paper, we propose a contour tree neural network for 3D surface segmentation in the context of hydrological applications. The model represents a 3D elevation surface by a contour tree skeleton from computational topology. Graph convolution and pooling operations are used in a multi-level hierarchy to learn node features at multiple surface scales. The unique advantage of the model compared with other existing models is that it can capture the topological structure of class segments along with surface contour patterns. Evaluations on real-world datasets show that the proposed model outperformed several baseline methods in classification accuracy.

One limitation in our current model is that the graph convolution filters assume specific contour tree structures, which often vary from one surface to another. Another limitation is that our model still relies on pre-extracted spatial contextual features from U-Net for each tree node due to the relatively small parameters in a graph neural network. We plan to address these issues in future work. For example, we can potentially study how to partition a geographic area along a river into pieces such that their contour tree structure is generally similar. We can also explore strategies to learn a U-Net model together with our CTNN in an end-to-end manner.

\newpage
\section*{Broader Impact}
This work has a significant potential societal impact in the flood mapping application.  Flood extent mapping plays a crucial role in addressing grand societal challenges such as disaster management,  national water forecasting, as well as energy and food security. For example, during Hurricane Harvey floods in 2017, first responders needed to know where flood water was in order to plan rescue efforts. In national water forecasting, detailed flood extent maps can be used to calibrate and validate the NOAA National Water Model~\cite{nwm}, which can forecast the flow of over 2.7 million rivers and streams through the entire continental U.S.~\cite{iwrss}. 
In current practice, flood extent maps are mostly generated by flood forecasting models, whose accuracy is often unsatisfactory in high spatial details~\cite{iwrss}. Other ways to generate flood maps involve sending a field crew on the ground to record high-water marks, or visually interpreting earth observation imagery~\cite{brivio2002integration}. However, the process is both expensive and time-consuming. With a large amount of high-resolution earth imagery being collected from satellites (e.g., DigitalGlobe, Planet Labs), aerial planes (e.g., NOAA National Geodetic Survey), and unmanned aerial vehicles, the cost of manually labeling flood extent becomes prohibitive.
This paper develops a classification model that can automatically classify earth observation imagery pixels into flood extent maps. The results can be used by first responders to plan rescue efforts, by hydrologists to calibrate and validate water forecasting models, as well as by insurance companies to process claims. The proposed model advances the state of the art methods by modeling the surface contour structures on geographic terrains. 

There are also some potential risks. First, the proposed graph neural network model is learned based on particular contour tree structures in the training area. If the contour tree topological skeleton varies from the training area to the test area, the model performance can degrade. This can impact model generalizability. One potential solution is to find ways to partition the surface more strategically, e.g., partitioning the area into different chunks along the river so that the contour tree skeletons from different chunks look similar. Another potential risk is the model interpretability. Our model is based on deep neural networks, which are black-box models in general and hard to interpret by a decision-making agency in real-world disaster response. This issue can potentially be mitigated by visualizing the predicted classes on surface contours in a 3D map through GIS software (e.g., ArcScene).



\bibliographystyle{unsrt}
{\small \bibliography{ref}}

\begin{thebibliography}{10}

\bibitem{natarajan2006segmenting}
Vijay Natarajan, Yusu Wang, Peer-Timo Bremer, Valerio Pascucci, and Bernd
  Hamann.
\newblock Segmenting molecular surfaces.
\newblock {\em Computer Aided Geometric Design}, 23(6):495--509, 2006.

\bibitem{shamir2008survey}
Ariel Shamir.
\newblock {A survey on mesh segmentation techniques}.
\newblock In {\em Computer graphics forum}, volume 27 (6), pages 1539--1556.
  Wiley Online Library, 2008.

\bibitem{nguyen20133d}
Anh Nguyen and Bac Le.
\newblock {3D point cloud segmentation: A survey}.
\newblock In {\em 2013 6th IEEE conference on robotics, automation and
  mechatronics (RAM)}, pages 225--230. IEEE, 2013.

\bibitem{kalvin1996superfaces}
Alan~D Kalvin and Russell~H Taylor.
\newblock Superfaces: Polygonal mesh simplification with bounded error.
\newblock {\em IEEE Computer Graphics and Applications}, 16(3):64--77, 1996.

\bibitem{grau2004improved}
Vicente Grau, AUJ Mewes, M~Alcaniz, Ron Kikinis, and Simon~K Warfield.
\newblock Improved watershed transform for medical image segmentation using
  prior information.
\newblock {\em IEEE transactions on medical imaging}, 23(4):447--458, 2004.

\bibitem{garland2001hierarchical}
Michael Garland, Andrew Willmott, and Paul~S Heckbert.
\newblock Hierarchical face clustering on polygonal surfaces.
\newblock In {\em Proceedings of the 2001 symposium on Interactive 3D
  graphics}, pages 49--58, 2001.

\bibitem{chen2012three}
Xinjian Chen, Meindert Niemeijer, Li~Zhang, Kyungmoo Lee, Michael~D
  Abr{\`a}moff, and Milan Sonka.
\newblock Three-dimensional segmentation of fluid-associated abnormalities in
  retinal oct: probability constrained graph-search-graph-cut.
\newblock {\em IEEE transactions on medical imaging}, 31(8):1521--1531, 2012.

\bibitem{minaee2020image}
Shervin Minaee, Yuri Boykov, Fatih Porikli, Antonio Plaza, Nasser Kehtarnavaz,
  and Demetri Terzopoulos.
\newblock Image segmentation using deep learning: A survey.
\newblock {\em arXiv preprint arXiv:2001.05566}, 2020.

\bibitem{ronneberger2015u}
Olaf Ronneberger, Philipp Fischer, and Thomas Brox.
\newblock U-net: Convolutional networks for biomedical image segmentation.
\newblock In {\em International Conference on Medical image computing and
  computer-assisted intervention}, pages 234--241. Springer, 2015.

\bibitem{long2015fully}
Jonathan Long, Evan Shelhamer, and Trevor Darrell.
\newblock Fully convolutional networks for semantic segmentation.
\newblock In {\em Proceedings of the IEEE conference on computer vision and
  pattern recognition}, pages 3431--3440, 2015.

\bibitem{chen2017deeplab}
Liang-Chieh Chen, George Papandreou, Iasonas Kokkinos, Kevin Murphy, and Alan~L
  Yuille.
\newblock Deeplab: Semantic image segmentation with deep convolutional nets,
  atrous convolution, and fully connected crfs.
\newblock {\em IEEE transactions on pattern analysis and machine intelligence},
  40(4):834--848, 2017.

\bibitem{chen2017rethinking}
Liang-Chieh Chen, George Papandreou, Florian Schroff, and Hartwig Adam.
\newblock Rethinking atrous convolution for semantic image segmentation.
\newblock {\em arXiv preprint arXiv:1706.05587}, 2017.

\bibitem{noh2015learning}
Hyeonwoo Noh, Seunghoon Hong, and Bohyung Han.
\newblock Learning deconvolution network for semantic segmentation.
\newblock In {\em Proceedings of the IEEE international conference on computer
  vision}, pages 1520--1528, 2015.

\bibitem{wang2016learning}
Jinghua Wang, Zhenhua Wang, Dacheng Tao, Simon See, and Gang Wang.
\newblock Learning common and specific features for rgb-d semantic segmentation
  with deconvolutional networks.
\newblock In {\em European Conference on Computer Vision}, pages 664--679.
  Springer, 2016.

\bibitem{ma2017multi}
Lingni Ma, J{\"o}rg St{\"u}ckler, Christian Kerl, and Daniel Cremers.
\newblock Multi-view deep learning for consistent semantic mapping with rgb-d
  cameras.
\newblock In {\em 2017 IEEE/RSJ International Conference on Intelligent Robots
  and Systems (IROS)}, pages 598--605. IEEE, 2017.

\bibitem{hazirbas2016fusenet}
Caner Hazirbas, Lingni Ma, Csaba Domokos, and Daniel Cremers.
\newblock Fusenet: Incorporating depth into semantic segmentation via
  fusion-based cnn architecture.
\newblock In {\em Asian conference on computer vision}, pages 213--228.
  Springer, 2016.

\bibitem{barron2016fast}
Jonathan~T Barron and Ben Poole.
\newblock The fast bilateral solver.
\newblock In {\em European Conference on Computer Vision}, pages 617--632.
  Springer, 2016.

\bibitem{wang2018depth}
Weiyue Wang and Ulrich Neumann.
\newblock Depth-aware cnn for rgb-d segmentation.
\newblock In {\em Proceedings of the European Conference on Computer Vision
  (ECCV)}, pages 135--150, 2018.

\bibitem{vicente2008graph}
Sara Vicente, Vladimir Kolmogorov, and Carsten Rother.
\newblock Graph cut based image segmentation with connectivity priors.
\newblock In {\em 2008 IEEE conference on computer vision and pattern
  recognition}, pages 1--8. IEEE, 2008.

\bibitem{nowozin2009global}
Sebastian Nowozin and Christoph~H Lampert.
\newblock Global connectivity potentials for random field models.
\newblock In {\em 2009 IEEE Conference on Computer Vision and Pattern
  Recognition}, pages 818--825. IEEE, 2009.

\bibitem{chen2011enforcing}
Chao Chen, Daniel Freedman, and Christoph~H Lampert.
\newblock Enforcing topological constraints in random field image segmentation.
\newblock In {\em CVPR 2011}, pages 2089--2096. IEEE, 2011.

\bibitem{andres2011probabilistic}
Bjoern Andres, J{\"o}rg~H Kappes, Thorsten Beier, Ullrich K{\"o}the, and Fred~A
  Hamprecht.
\newblock Probabilistic image segmentation with closedness constraints.
\newblock In {\em 2011 International Conference on Computer Vision}, pages
  2611--2618. IEEE, 2011.

\bibitem{hu2019topology}
Xiaoling Hu, Fuxin Li, Dimitris Samaras, and Chao Chen.
\newblock Topology-preserving deep image segmentation.
\newblock In {\em Advances in Neural Information Processing Systems}, pages
  5658--5669, 2019.

\bibitem{mosinska2018beyond}
Agata Mosinska, Pablo Marquez-Neila, Mateusz Kozi{\'n}ski, and Pascal Fua.
\newblock Beyond the pixel-wise loss for topology-aware delineation.
\newblock In {\em Proceedings of the IEEE Conference on Computer Vision and
  Pattern Recognition}, pages 3136--3145, 2018.

\bibitem{he2019deep}
Yufan He, Aaron Carass, Yihao Liu, Bruno~M Jedynak, Sharon~D Solomon, Shiv
  Saidha, Peter~A Calabresi, and Jerry~L Prince.
\newblock Deep learning based topology guaranteed surface and mme segmentation
  of multiple sclerosis subjects from retinal oct.
\newblock {\em Biomedical optics express}, 10(10):5042--5058, 2019.

\bibitem{lang2017improving}
Andrew Lang, Aaron Carass, Ava~K Bittner, Howard~S Ying, and Jerry~L Prince.
\newblock Improving graph-based oct segmentation for severe pathology in
  retinitis pigmentosa patients.
\newblock In {\em Medical Imaging 2017: Biomedical Applications in Molecular,
  Structural, and Functional Imaging}, volume 10137, page 101371M.
  International Society for Optics and Photonics, 2017.

\bibitem{edelsbrunner2010computational}
Herbert Edelsbrunner and John Harer.
\newblock {\em Computational topology: an introduction}.
\newblock American Mathematical Soc., 2010.

\bibitem{munkres2000topology}
James~R Munkres.
\newblock {\em Topology}.
\newblock Pearson, January 2000.

\bibitem{carr2003computing}
Hamish Carr, Jack Snoeyink, and Ulrike Axen.
\newblock Computing contour trees in all dimensions.
\newblock {\em Computational Geometry}, 24(2):75--94, 2003.

\bibitem{zhou2018graph}
Jie Zhou, Ganqu Cui, Zhengyan Zhang, Cheng Yang, Zhiyuan Liu, Lifeng Wang,
  Changcheng Li, and Maosong Sun.
\newblock Graph neural networks: A review of methods and applications.
\newblock {\em arXiv preprint arXiv:1812.08434}, 2018.

\bibitem{zhang2018deep}
Ziwei Zhang, Peng Cui, and Wenwu Zhu.
\newblock Deep learning on graphs: A survey.
\newblock {\em arXiv preprint arXiv:1812.04202}, 2018.

\bibitem{wu2019comprehensive}
Zonghan Wu, Shirui Pan, Fengwen Chen, Guodong Long, Chengqi Zhang, and Philip~S
  Yu.
\newblock A comprehensive survey on graph neural networks.
\newblock {\em arXiv preprint arXiv:1901.00596}, 2019.

\bibitem{chiang2019cluster}
Wei-Lin Chiang, Xuanqing Liu, Si~Si, Yang Li, Samy Bengio, and Cho-Jui Hsieh.
\newblock Cluster-gcn: An efficient algorithm for training deep and large graph
  convolutional networks.
\newblock In {\em Proceedings of the 25th ACM SIGKDD International Conference
  on Knowledge Discovery \& Data Mining}, pages 257--266, 2019.

\bibitem{gao2018large}
Hongyang Gao, Zhengyang Wang, and Shuiwang Ji.
\newblock Large-scale learnable graph convolutional networks.
\newblock In {\em Proceedings of the 24th ACM SIGKDD International Conference
  on Knowledge Discovery \& Data Mining}, pages 1416--1424, 2018.

\bibitem{defferrard2016convolutional}
Micha{\"e}l Defferrard, Xavier Bresson, and Pierre Vandergheynst.
\newblock Convolutional neural networks on graphs with fast localized spectral
  filtering.
\newblock In {\em Advances in neural information processing systems}, pages
  3844--3852, 2016.

\bibitem{li2017diffusion}
Yaguang Li, Rose Yu, Cyrus Shahabi, and Yan Liu.
\newblock Diffusion convolutional recurrent neural network: Data-driven traffic
  forecasting.
\newblock {\em arXiv preprint arXiv:1707.01926}, 2017.

\bibitem{dhillon2007weighted}
Inderjit~S Dhillon, Yuqiang Guan, and Brian Kulis.
\newblock Weighted graph cuts without eigenvectors a multilevel approach.
\newblock {\em IEEE transactions on pattern analysis and machine intelligence},
  29(11):1944--1957, 2007.

\bibitem{ying2018hierarchical}
Zhitao Ying, Jiaxuan You, Christopher Morris, Xiang Ren, Will Hamilton, and
  Jure Leskovec.
\newblock Hierarchical graph representation learning with differentiable
  pooling.
\newblock In {\em Advances in neural information processing systems}, pages
  4800--4810, 2018.

\bibitem{gao2019graph}
Hongyang Gao and Shuiwang Ji.
\newblock Graph u-nets.
\newblock {\em arXiv preprint arXiv:1905.05178}, 2019.

\bibitem{ngs}
{National Oceanic and Atmospheric Administration}.
\newblock Data and imagery from noaa's national geodetic survey.
\newblock \url{https://www.ngs.noaa.gov}, 2017.

\bibitem{ncsudem}
{NCSU Libraries}.
\newblock {LIDAR Based Elevation Data for North Carolina}.
\newblock \url{https://www.lib.ncsu.edu/gis/elevation}, 2018.

\bibitem{nwm}
{National Oceanic and Atmospheric Administration}.
\newblock {National Water Model: Improving NOAA's Water Prediction Services}.
\newblock \url{http://water.noaa.gov/documents/wrn-national-water-model.pdf},
  2018.

\bibitem{iwrss}
Don Cline.
\newblock Integrated water resources science and services: an integrated and
  adaptive roadmap for operational implementation.
\newblock Technical report, National Oceanic and Atmospheric Administration,
  2009.

\bibitem{brivio2002integration}
PA~Brivio, R~Colombo, M~Maggi, and R~Tomasoni.
\newblock Integration of remote sensing data and gis for accurate mapping of
  flooded areas.
\newblock {\em International Journal of Remote Sensing}, 23(3):429--441, 2002.

\end{thebibliography}
\clearpage
\appendix

\section{Implementation Details}

\subsection{U-Net model}

The U-Net model architecture is shown in Table~\ref{tab:unet}. The model consists of five double convolutional layers and max-pooling layers in the downsample path as well as five double convolutional layers and transposed convolutional layers in the upsample path. There is a batch normalization operation within each convolutional layer before non-linear activation based on ReLU (rectified linear unit). The model has 31,455,042 trainable parameters in total. 
\begin{table}[h]\footnotesize
\centering
\caption{U-Net model architecture}
\label{tab:unet}
\begin{tabular}{ccc}
\toprule
Layer (type) & Output shape & Param \# \\ \hline
Input &(None, 224, 224, 4)&0\\ \hline  
Conv2D &(None, 224, 224, 32)&1184\\ \hline  
Conv2D&(None, 224, 224, 32)&9248\\ \hline  
Max pooling&(None, 112, 112, 32)&0\\ \hline 
Conv2D&(None, 112, 112, 64)&18496\\ \hline

Conv2D&(None, 112, 112, 64)&36928\\ \hline  
Max pooling&(None, 56, 56, 64)&0\\ \hline  
Conv2D&(None, 56, 56, 128)&73856\\ \hline  
 Conv2D&(None, 56, 56, 128)&147584\\ \hline
Max pooling&(None, 28, 28, 128)&0\\ \hline
  
 Conv2D&(None, 28, 28, 256)&295168\\ \hline  
 Conv2D&(None, 28, 28, 256)&590080\\ \hline
Max pooling&(None, 14, 14, 256)&0\\ \hline

 Conv2D&(None, 14, 14, 512)&1180160\\ \hline  
 Conv2D&(None, 14, 14, 512)&2359808\\ \hline
Max pooling&(None, 7, 7, 512)&0\\ \hline

 Conv2D&(None, 7, 7, 1024)&4719616\\ \hline  
 Conv2D&(None, 7, 7, 1024)&9438208\\ \hline
Up sampling&(None, 14, 14, 1024)&0\\ \hline
 Conv2D&(None, 14, 14, 512)&7078400\\ \hline  
 Conv2D&(None, 14, 14, 512)&2359808\\ \hline
Up sampling&(None, 28, 28, 512)&0\\ \hline

 Conv2D&(None, 28, 28, 256)&1769728\\ \hline  
 Conv2D&(None, 28, 28, 256)&590080\\ \hline
Up sampling&(None, 56, 56, 256)&0\\ \hline
 Conv2D&(None, 56, 56, 128)&442496\\ \hline  
 Conv2D&(None, 56, 56, 128)&147584\\ \hline
Up sampling&(None, 112, 112, 128)&0\\ \hline

 Conv2D&(None, 112, 112, 64)&110656\\ \hline  
 Conv2D&(None, 112, 112, 64)&36928\\ \hline
Up sampling&(None, 224, 224, 64)&0\\ \hline

Conv2D&(None, 224, 224, 32)&27680\\ \hline  
 Conv2D&(None, 224, 224, 32)&9248\\ \bottomrule

\end{tabular}
\end{table}

\subsection{Hyper-parameters}

{\bf The CTNN model} has several hyperparameters, including the number of hops $h$, the number of output channels $F_g$ in ChebyNet or diffusion graph convolution layer, the number of downsampling levels $L$ together with the precision of elevation values for constructing contour tree at each level ($r_0,r_1,...,r_L$).  The original precision of elevation $r_0$ is 0.01, then we keep reducing the precision to $r_1, ..., r_L$. The number of downsampling levels $L$ indicates the number of max-pooling operations in the model architecture.
Therefore, the number of graph convolutional levels is $L+1$ in the CTNN model.

\begin{table}[h]\footnotesize
  \centering
  \caption{Hyperparameters of CTNN in two flood datasets}
  \label{tab:hpctnn}
  \begin{tabular}{cl}
    \toprule
    Hyperparameter&Value\\
    \midrule
    $L$&3 \\
    $h$ &[4, 4, 2, 2] \\
    $F_g$&[16, 32, 64, 128] \\
    
    $r_l (\text{meter})$&[0.01, 0.1, 1, 10] \\
  \bottomrule
\end{tabular}
\end{table}

\subsection{Hyper-parameter tuning}
The selection of hyperparameter values was based on the exploration of data characteristics and the evaluation of validation data. For example, for elevation values, we chose the precision with 3 levels, from 0.01-meter to 0.1-meter, 1 meter, 10-meter. For the Grimland dataset,  the original contour tree has around 50,000 nodes at 0.01-meter precision. The collapsed contour tree has around 8,000 nodes in 0.1-meter precision, 1,000 nodes in 1-meter precision, and 100 nodes in 10-meter precision. Different image patches may have a slightly different number of tree nodes in each spatial resolution.  These levels helped simplify the topology of the contour tree from a fine resolution (with local topology) to a coarse resolution (with global topology trend). The number of hops and output channels can be somehow determined by the validation dataset. 
\subsection{Model training}
{\bf For U-Net model} , we applied the model architecture in Table~\ref{tab:unet} with Keras. In model training, we used Adam optimizer. The loss function was binary cross-entropy. The learning rate was $10^{-4}$. 

\begin{figure}[h]
    \centering
    \vspace{-5mm}
    \subfloat[Grimesland dataset] {\includegraphics[height=1.5in]{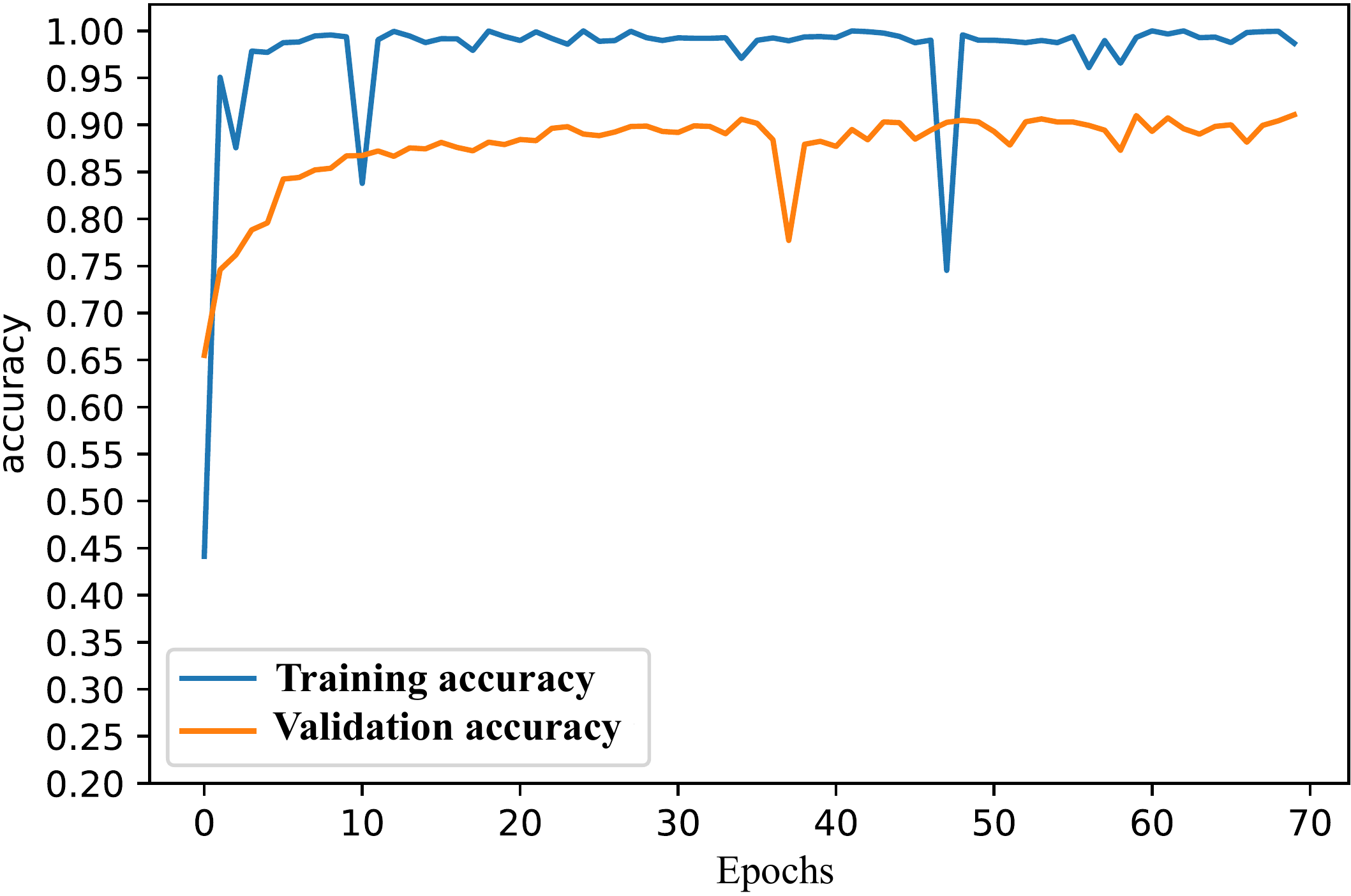}}
    \hspace{4mm}
    \subfloat[Kinston dataset] {\includegraphics[height=1.5in]{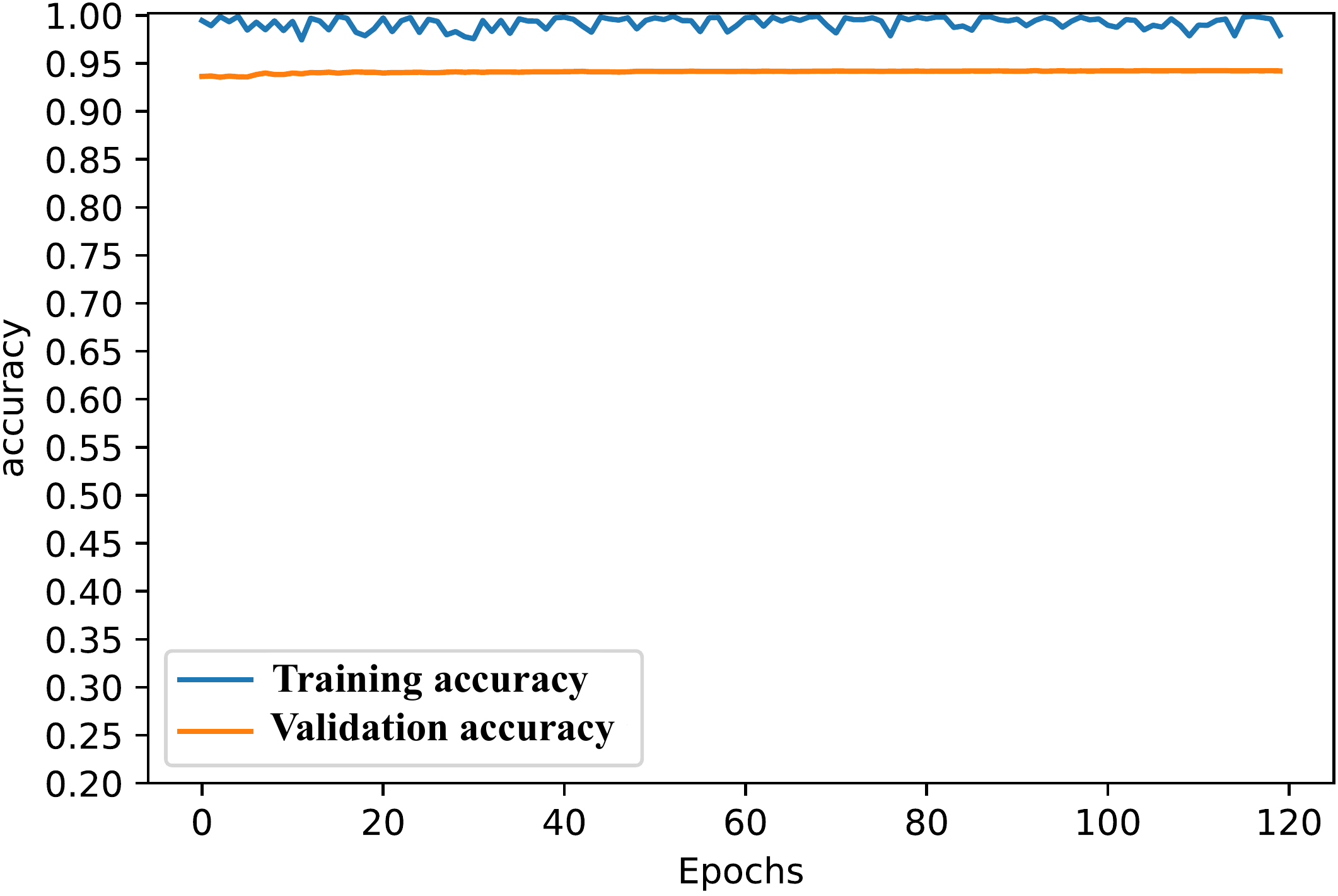}}  
    \caption{The training curve of CTNN model of Grimesland dataset (a), and Kinston dataset (b)}
    \vspace{-5mm}
    \label{fig:tr}
\end{figure}

{\bf For CTNN model}, we used the loss function called {\em sparse softmax cross-entropy with logits} in Tensorflow. We used the MomentumOptimizer optimizer with a momentum of 0.9, a learning rate of $10^{-4}$, a decaying rate of 0.99, and a $L_2$ regularization weight of $5\times10^{-2}$. We also added a batch normalization layer before the non-linear activation in each graph convolution layer. We trained the model for 70 epochs with a mini-batch size of $1$, since the graph topology of the input patches was not fixed. Figure~\ref{fig:tr}  show the training curve of the CTNN model of two datasets. 

\subsection{Prediction results}
\begin{figure}[h]
    \centering
    \vspace{-5mm}
    \subfloat[Aerial Image] {\includegraphics[height=1.5in]{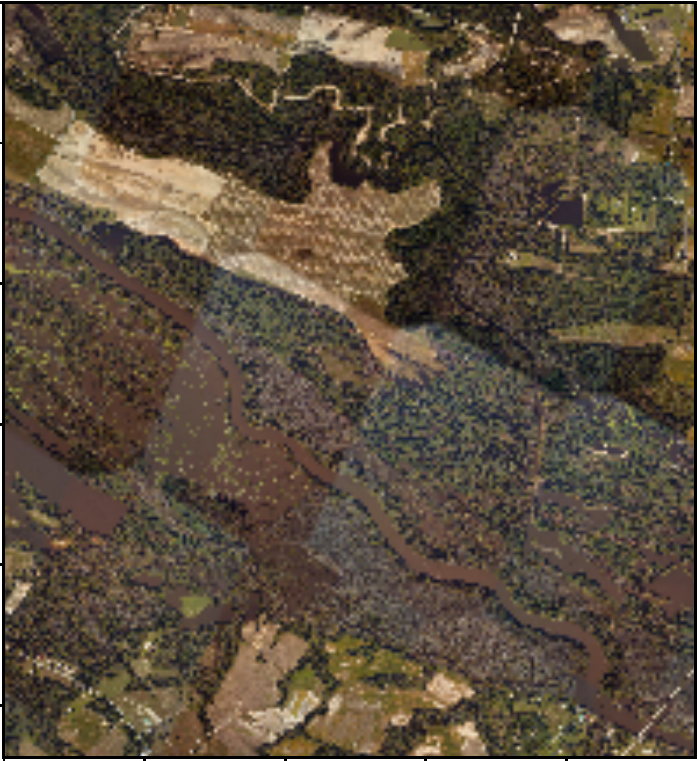}}
    \hspace{4mm}
    \subfloat[Digital Elevation] {\includegraphics[height=1.5in]{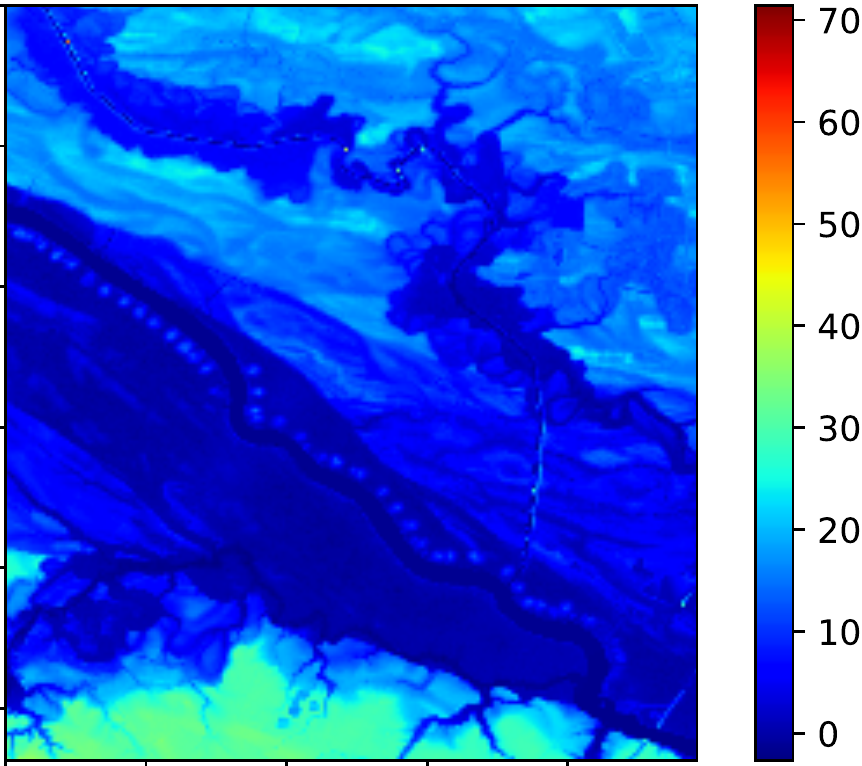}}  
    \hspace{4mm}
    \subfloat[Ground Truth] {\includegraphics[height=1.5in]{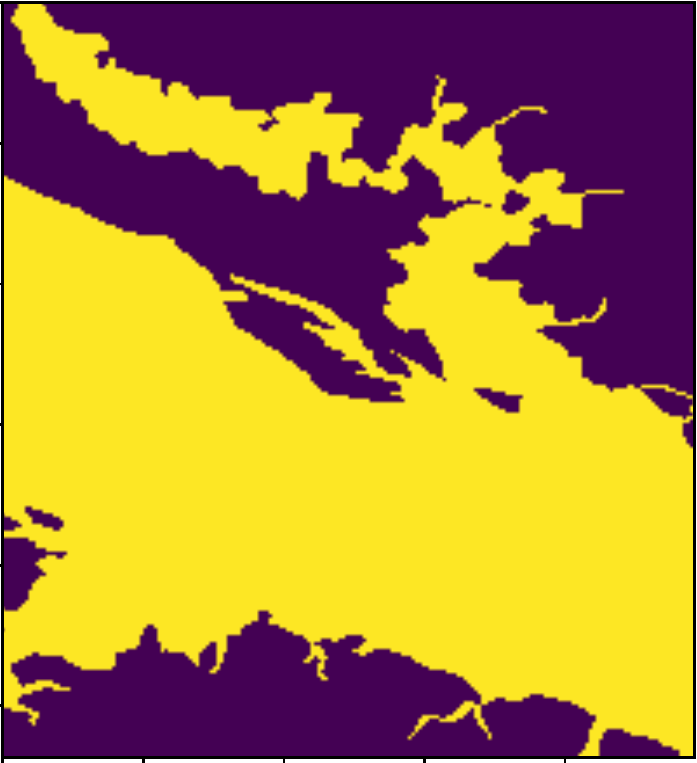}}  
    \hspace{4mm}
    \subfloat[Prediction  of CTNN model] {\includegraphics[height=1.5in]{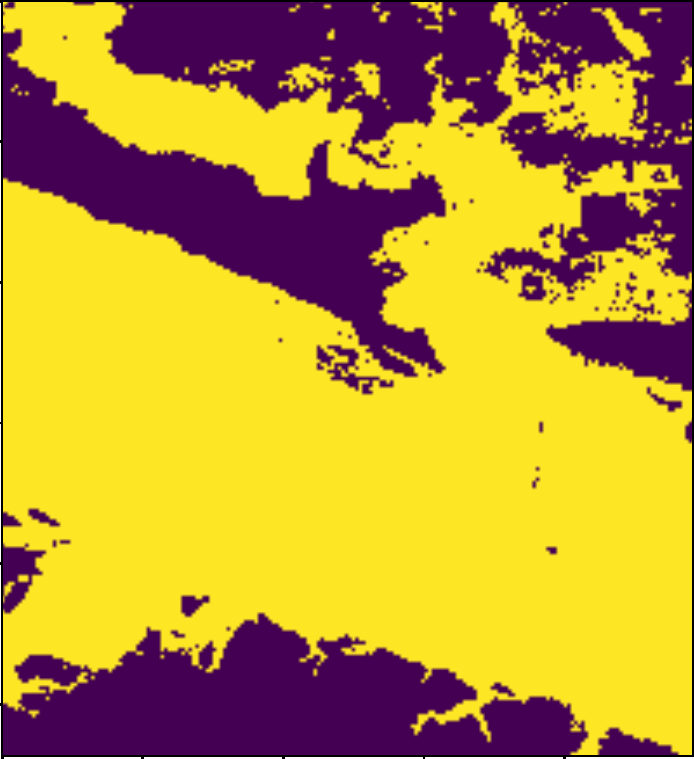}}  
    \caption{The aerial image (a), digital elevation (b), ground truth (c), and CTNN model prediction  of Grimesland dataset  (Yellow is flood class, purple is dry class)}
    \vspace{-5mm}
    \label{fig:im1}
\end{figure}

\begin{figure}[h]
    \centering
    \vspace{-5mm}
    \subfloat[Aerial Image] {\includegraphics[height=1.5in]{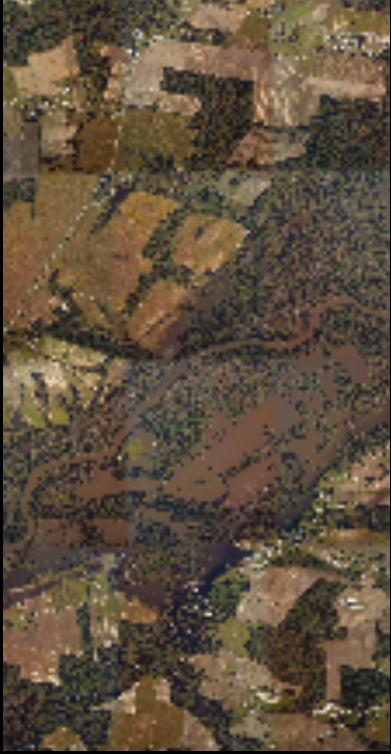}}
    \hspace{4mm}
    \subfloat[Digital Elevation] {\includegraphics[height=1.5in]{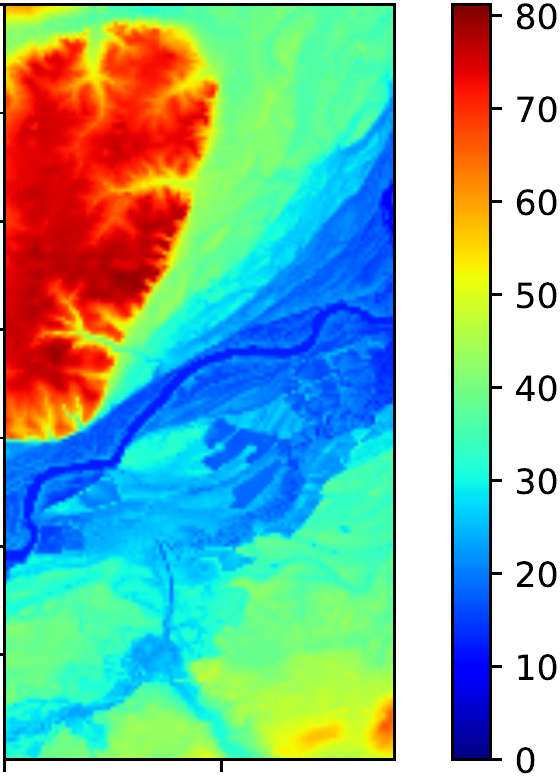}}  
    \hspace{4mm}
    \subfloat[Ground Truth] {\includegraphics[height=1.5in]{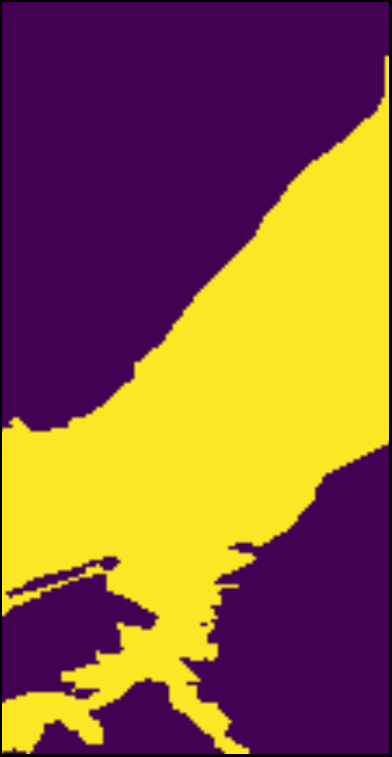}}  
    \hspace{4mm}
    \subfloat[Prediction  of CTNN model] {\includegraphics[height=1.5in]{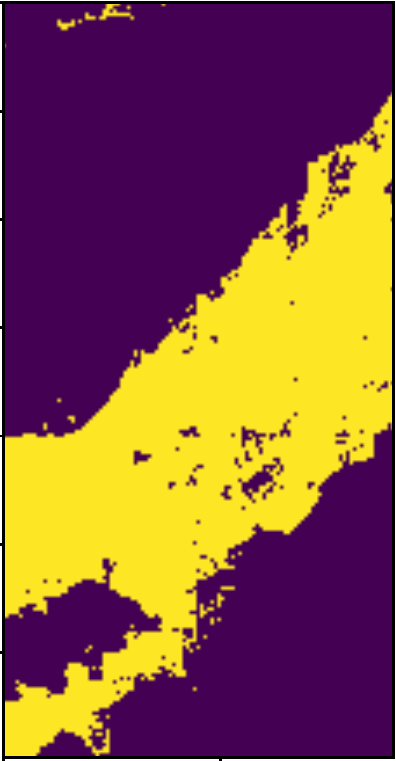}}  
    \caption{The aerial image (a), digital elevation (b), ground truth (c), and CTNN model prediction  of Kinston dataset (Yellow is flood class, purple is dry class)}
    \vspace{-5mm}
    \label{fig:im2}
\end{figure}

In Figure~\ref{fig:im1} and Figure~\ref{fig:im2}, we provide the aerial image, digital elevation, and ground truth of two datasets together with the prediction of the CTNN model on two datasets. For the Grimesland dataset, the false-positive rate is high because of the feature ambiguity on the upper right corner of the test image. For the Kinston dataset, the false-negative rate is high because of the noise in the middle part of the test image. 

\end{document}